\documentclass[pdflatex,sn-mathphys-num]{sn-jnl}                  
\usepackage{microtype}
\usepackage{graphicx}
\usepackage{caption}
\usepackage{subcaption}
\usepackage{adjustbox}
\usepackage{longtable}
\usepackage{multirow}
\usepackage{makecell}
\usepackage{algorithmicx}
\usepackage[ruled]{algorithm}
\usepackage{algpseudocode}
\usepackage[table,xcdraw]{xcolor}
\usepackage{float}
\usepackage{amssymb}
\usepackage{bbm}
\usepackage{xcolor}
\usepackage{soul}
\usepackage{amsmath}
\usepackage{hyperref}
\usepackage{booktabs}
\usepackage{tikz}
\usepackage[T1]{fontenc}
\usetikzlibrary{positioning, arrows.meta}
                                    
\theoremstyle{thmstyleone}

\theoremstyle{thmstyletwo}

\theoremstyle{thmstylethree}

\begin{document}

\title[Article Title]{UCTECG-Net: Uncertainty-aware Convolution Transformer ECG Network for Arrhythmia Detection}

\author*[1]{\fnm{Hamzeh} \sur{Asgharnezhad}}\email{h.asgharnezhad@deakin.edu.au}

\author[1]{\fnm{Pegah} \sur{Tabarisaadi}}

\author[1]{\fnm{Abbas} \sur{Khosravi}}\email{abbas.khosravi@deakin.edu.au}

\author[1]{\fnm{Roohallah} \sur{Alizadehsani}}

\author[2]{\fnm{U. Rajendra} \sur{Acharya}}

\affil*[1]{%
\orgname{Deakin University}, \country{Australia}
}

\affil[2]{%
\orgdiv{School of Mathematics, Physics and Computing},
\orgname{University of Southern Queensland},
\orgaddress{\city{Toowoomba}, \state{QLD}, \country{Australia}}
}


\abstract{
Deep learning has improved automated electrocardiogram (ECG) classification, but limited insight into prediction reliability hinders its use in safety-critical settings. This paper proposes UCTECG-Net, an uncertainty-aware hybrid architecture that combines one-dimensional convolutions and Transformer encoders to process raw ECG signals and their spectrograms jointly. Evaluated on the MIT-BIH Arrhythmia and PTB Diagnostic datasets, UCTECG-Net outperforms LSTM, CNN1D, and Transformer baselines in terms of accuracy, precision, recall and F1 score, achieving up to 98.58\% accuracy on MIT-BIH and 99.14\% on PTB.
To assess predictive reliability, we integrate three uncertainty quantification methods (Monte Carlo Dropout, Deep Ensembles, and Ensemble Monte Carlo Dropout) into all models and analyze their behavior using an uncertainty-aware confusion matrix and derived metrics. The results show that UCTECG-Net, particularly with Ensemble or EMCD, provides more reliable and better-aligned uncertainty estimates than competing architectures, offering a stronger basis for risk-aware ECG decision support.
}


\maketitle


\section{Introduction}
In recent years, artificial intelligence (AI) and machine learning (ML) have revolutionized numerous domains by enabling systems to learn from data, identify patterns, and make data-driven decisions with minimal human intervention \cite{hao2022physics}. These technologies have demonstrated remarkable success across diverse fields, including finance, transportation, cyber security, and, notably, healthcare. In the medical domain, ML has unlocked powerful new possibilities for automating diagnosis, improving clinical workflows, and enabling personalized medicine. By training models on large-scale clinical datasets, researchers have demonstrated the ability of ML to assist in detecting various conditions such as skin cancer from dermoscopic images \cite{asgharnezhad2025quadstepapproachuncertaintyawaredeep, tabarisaadi2022uncertainty}, arrhythmias from ECG signals \cite{hannun2019cardiologist}, pneumonia and tuberculosis from chest X-rays \cite{rajpurkar2017chexnet}, and diabetic retinopathy from retinal scans \cite{gulshan2016development}. These tools offer substantial benefits by helping clinicians interpret complex data, prioritize patient care, and reduce time-consuming manual tasks, ultimately promoting more scalable and cost-effective healthcare delivery.

Despite these advantages, the widespread adoption of ML in clinical settings faces critical challenges related to trust, reliability, and interoperability \cite{amann2020explainability}. Many high-performing ML models, particularly deep neural networks, operate as “black boxes,” offering little transparency into how decisions are made \cite{rudin2019stop}. This opacity makes it difficult to assess when and why a model might fail a grave concern in high-stakes environments like healthcare, where errors can have life-threatening consequences \cite{cabitza2017unintended, watson2019clinical}. Failures can occur due to distribution shifts, noisy or insufficient data, or unseen patient profiles. To address these challenges, recent research has emphasized the importance of uncertainty quantification (UQ), which enables models not only to provide predictions but also to express confidence in their outputs\cite{tabarisaadi2022optimized}. UQ allows systems to distinguish between reliable and unreliable predictions, flagging cases where caution is warranted and human oversight is necessary. This helps build trust in automated systems, supports risk-aware decision-making, and enhances patient safety. Several methods have been proposed for UQ in deep learning, including Monte Carlo dropout \cite{gal2016dropout}, deep ensembles \cite{lakshminarayanan2017simple}, and Bayesian neural networks \cite{kendall2017uncertainties}, which have been successfully applied in both medical image analysis and physiological signal classification. As such, integrating UQ into ML models is a critical step toward making AI-driven diagnosis both safer and more clinically useful.

ML techniques have been extensively applied to electrocardiogram (ECG) signals to enable automated interpretation, particularly for detecting cardiac abnormalities such as arrhythmias, atrial fibrillation, and myocardial infarction \cite{hannun2019cardiologist, faust2018deep, attia2019artificial}. 
Traditional approaches typically involved handcrafted feature extraction—such as QRS complex duration, RR intervals, and heart rate variability—followed by classification with algorithms including support vector machine (SVM), k-nearest neighbors (k-NN), or random forests \cite{osowski2001ecg, ince2009generic}. While these methods achieved reasonable performance, they often relied heavily on expert-designed features and were sensitive to noise or variability in ECG morphology. In recent years, deep learning models have gained popularity due to their ability to learn discriminative features directly from raw ECG waveforms. Convolutional neural networks (CNNs) have been successfully employed to capture local temporal patterns \cite{oh2018automated}, while recurrent neural networks (RNNs), particularly long short-term memory (LSTM) units, have been used to model the sequential nature of ECG signals \cite{yildirim2018novel}. More advanced architectures, including transformer-based models \cite{che2021constrained} and hybrid CNN-LSTM networks [9], have also demonstrated strong performance in large-scale ECG classification tasks. These deep models have been trained on datasets such as the MIT-BIH Arrhythmia Database and the PhysioNet Challenge datasets \cite{moody2001impact, goldberger2000physiobank}, enabling the detection of multiple heartbeat types and rhythms with high granularity. As a result, machine learning has significantly advanced the automation and reliability of ECG interpretation, facilitating faster diagnosis and continuous patient monitoring \cite{minchole2019artificial}.

Recently, several studies have explored transforming ECG signals into spectrogram representations to better capture their non-stationary and time–frequency characteristics. Spectrograms convert the one-dimensional ECG waveform into a two–dimensional time–frequency map, revealing how signal energy evolves across different frequency bands. This representation allows deep learning models, particularly convolutional neural networks (CNNs), to extract discriminative spatial–spectral features that are often hidden in the raw temporal domain. By leveraging these visual-like inputs, spectrogram-based models can utilize well-established architectures from computer vision and demonstrate enhanced robustness to noise, morphological variations, and inter-patient differences. Consequently, spectrogram-based deep learning approaches have achieved superior performance in detecting arrhythmias and other cardiac abnormalities compared with traditional signal-based methods \cite{acharya2017deep, zheng2020deep, kiranyaz2021ecg, wang2023spectrogram}.

UQ has recently gained attention in ECG analysis as a means to improve the reliability and trustworthiness of automated diagnostic systems. Given the critical nature of cardiovascular diagnosis and the potential consequences of misclassification, it is essential for machine learning models to not only predict outcomes but also express confidence in their predictions. In the context of ECG signal classification, UQ techniques help identify predictions that may be unreliable due to noisy inputs, distribution shifts, or ambiguous waveform patterns. Several studies have applied Bayesian deep learning approaches such as Monte Carlo (MC) dropout and deep ensembles to estimate predictive uncertainty in arrhythmia detection tasks \cite{abdar2021review}. These methods provide uncertainty scores alongside predictions, allowing clinicians to flag low-confidence cases for further human review \cite{pearce2020uncertainty}. For instance, Ghoshal and Tucker \cite{ghoshal2021estimating} applied MC dropout to ECG-based COVID-19 diagnosis and demonstrated that uncertainty estimates could effectively identify high-risk misclassifications. Additionally, research has shown that uncertainty-aware models can improve model calibration, reduce overconfidence, and enhance clinical decision-making when integrated with human-in-the-loop systems \cite{postels2019sampling, begoli2019need}. Despite these advances, uncertainty quantification in ECG remains an emerging area, and further research is needed to standardize evaluation metrics, improve the interpretability of uncertainty estimates, and ensure robustness in real-world clinical settings.

This paper advances uncertainty-aware ECG classification by addressing the limited focus on predictive reliability in existing deep learning approaches, which predominantly emphasize accuracy and often rely on either temporal or frequency-domain features in isolation. To systematically study this gap, we evaluate multiple deep learning architectures, including a 1D CNN, an LSTM, and a Transformer-based model, on widely used ECG Heartbeat Datasets using standard performance metrics such as accuracy, uncertainty accuracy, precision, recall and F1 score. Building on this analysis, we propose UCTECG-Net. This hybrid architecture jointly processes raw ECG signals and their spectrogram representations using convolutional and Transformer layers to capture both temporal and spectral characteristics. Unlike prior works, UCTECG-Net integrates multiple uncertainty quantification techniques within a unified framework, enhancing both classification accuracy and predictive reliability. To investigate uncertainty estimation in this context, we apply three prominent uncertainty quantification methods— Monte Carlo Dropout (MCD) \cite{gal2016dropout}, Deep Ensembles \cite{lakshminarayanan2017simple}, and a hybrid Ensemble–MCD approach—to the Transformer and UCTECG-Net models. We compare their predictive performance and uncertainty awareness to assess their ability to flag uncertain predictions and to support more reliable automated ECG interpretation. The findings provide valuable insights into the trade-offs between model accuracy and uncertainty calibration, highlighting the practical importance of integrating uncertainty-aware learning into real-world, safety-critical ECG diagnostic systems.

The main contributions of this paper are summarized as follows:
\begin{itemize}
    \item A systematic evaluation of multiple deep learning architectures (CNN1D, LSTM, and Transformer) is conducted for ECG heartbeat classification, enabling a unified performance comparison across widely used ECG datasets.
    \item A novel hybrid uncertainty-aware architecture, termed UCTECG-Net, is introduced, in which raw ECG waveforms and their corresponding spectrogram representations are processed through parallel convolutional and Transformer-based branches, enabling the joint learning of localized morphological patterns and long-range temporal--spectral dependencies.
    \item Multiple uncertainty quantification techniques, including Monte Carlo Dropout, Deep Ensembles, and a hybrid Ensemble--MCD approach, are integrated within a unified ECG classification framework.
    \item A comprehensive assessment of both classification performance and uncertainty-aware behavior is performed, demonstrating the effectiveness of uncertainty estimates in identifying unreliable predictions and supporting risk-aware decision-making.
    \item Empirical insights into the trade-offs between predictive accuracy and uncertainty calibration are provided, highlighting the practical importance of uncertainty-aware deep learning for trustworthy ECG diagnostic systems.
\end{itemize}

The remainder of this paper is structured as follows. Section~\ref{Background} reviews the Bayesian foundations of uncertainty estimation, introduces the uncertainty quantification techniques adopted in this work, and outlines the metrics used to evaluate both predictive performance and uncertainty quality. Section~\ref{sec:dataset} describes the MIT-BIH and PTB datasets, details the data partitioning and resampling procedures, and explains how the ECG signals and spectrograms are prepared for input to the models. Section~\ref{Simulation and Results} presents the experimental setup, reports the classification performance of UCTECG-Net and the baseline architectures, and analyzes their uncertainty-aware behavior. Finally, Section~\ref{conclusion} summarizes the key findings and concludes the paper.
\section{Background}\label{Background}
In supervised learning, the Bayesian framework provides a principled approach to quantify uncertainty in model predictions. A Bayesian neural network (BNN) models the posterior distribution over the weights $\mathbf{W}$ given the observed data $\mathcal{D}$, denoted as $p(\mathbf{W}|\mathcal{D})$. The predictive distribution for a new input $\mathbf{x}$ is then obtained by integrating over this posterior:

\begin{equation} p(y|\mathbf{x}, \mathcal{D}) = \int p(y|\mathbf{x}, \mathbf{W}) p(\mathbf{W}|\mathcal{D}) , d\mathbf{W} \end{equation}

Exact computation of this integral is generally intractable for deep neural networks due to the high dimensionality and complexity of the weight space. Therefore, approximate inference methods are employed to make Bayesian learning feasible in practice.
\subsection{Monte Carlo Dropout (MCD)}
One widely used Bayesian approximation technique is Monte Carlo Dropout. A key challenge in Bayesian neural networks is the intractability of the posterior distribution over model weights. Gal and Ghahramani \cite{gal2016dropout} proposed that this posterior can be approximated by applying dropout during inference and performing multiple stochastic forward passes through the network. This method allows for efficient sampling from the approximate posterior with minimal computational overhead.
For a given input $x$, the predictive mean over $T$ stochastic passes is calculated as:

\begin{equation} \mu_{pred} \approx \frac{1}{T} \sum_{t=1}^{T} p(y = c \mid x, \hat{\omega}_t) \end{equation}

where $p(y = c \mid x, \hat{\omega}_t)$ denotes the softmax probability of class $c$ given the sampled weights $\hat{\omega}_t$ from the $t$-th forward pass, and $T$ is the number of samples.

In addition to the predictive mean, the model's uncertainty can be estimated. Gal proposed predictive entropy as a measure of uncertainty \cite{gal2016dropout}:

\begin{equation} \label{Eq:MC-Dropout-PE} PE = - \sum_c \mu_{pred} \log \mu_{pred} \end{equation}

where the summation is over all possible output classes $c$. Predictive entropy reflects the dispersion of predicted class probabilities; low entropy implies a confident prediction, while high entropy indicates uncertainty. In classification tasks, this metric is useful for identifying uncertain or ambiguous predictions.

\begin{figure}[t]
    \centering
    \includegraphics[width=\columnwidth]{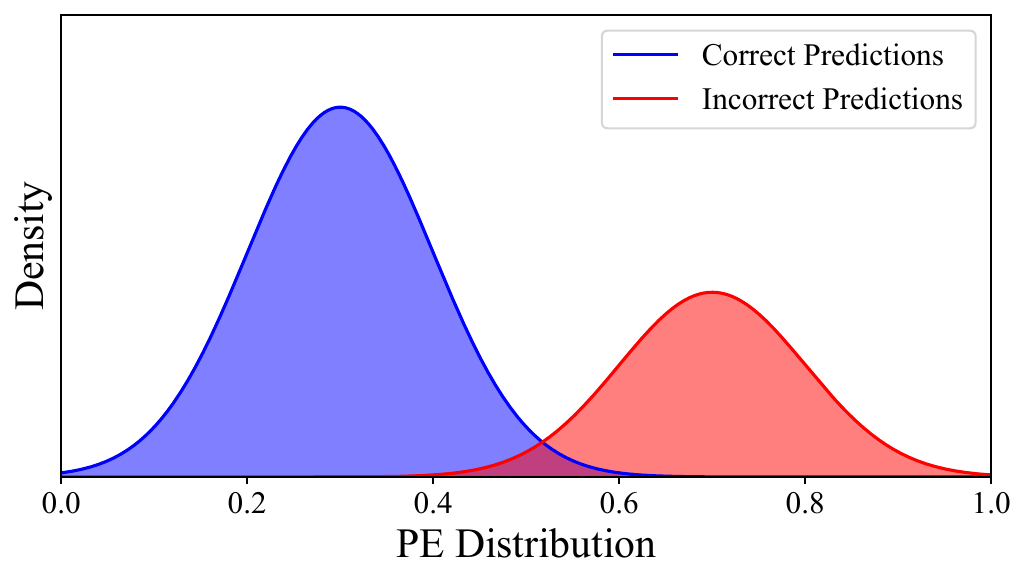}
    \caption{Density plots of certainty values for correct and incorrect predictions, illustrating the model’s ability to separate confident correct outputs from uncertain errors.}
    \label{fig:pe_distribution}
\end{figure}
\subsection{Deep Ensembles}
Deep ensembles offer a straightforward yet effective approach to uncertainty estimation by exploiting the variability among multiple independently trained neural networks. Rather than approximating a Bayesian posterior, this method involves training $N$ separate models with different random initializations, and optionally different data shuffling or bootstrapping strategies \cite{lakshminarayanan2017simple}. The diversity across these models enables the capture of epistemic uncertainty through variations in their predictions.

For a given test input $x$, each model in the ensemble produces a predictive distribution $p_{\theta_i}(y|x)$, where $\theta_i$ represents the parameters of the $i^{th}$ model. The ensemble-averaged predictive distribution is computed as:

\begin{equation} \hat{p}(y|x) = \frac{1}{N} \sum_{i=1}^{N} p_{\theta_i}(y|x) \end{equation}

The uncertainty associated with the ensemble’s prediction can be quantified using predictive entropy (PE), defined as:

\begin{equation} PE = -\sum_{i=0}^{C} \hat{p}(y_i|x) \log \hat{p}(y_i|x) \end{equation}

where $C$ is the total number of output classes. A lower entropy value indicates greater confidence in the prediction, whereas a higher value reflects increased uncertainty.

As demonstrated by Lakshminarayanan et al.\cite{lakshminarayanan2017simple}, deep ensembles can produce well-calibrated predictive distributions and effectively quantify uncertainty, all without requiring explicit Bayesian inference techniques.
\subsection{Ensemble Monte Carlo Dropout (EMCD)}
Ensemble Monte Carlo Dropout (EMCD) is a hybrid uncertainty estimation technique that integrates the advantages of deep ensembles and Monte Carlo Dropout. In this approach, $N$ neural networks are trained independently, similar to standard ensemble methods. During inference, each model performs $T$ stochastic forward passes with dropout enabled, following the procedure used in Monte Carlo Dropout.

For a given input $x$, the predictive distribution from the $i^{th}$ model, denoted by $\theta_i$, is computed by averaging the predictions from $T$ dropout-induced forward passes:

\begin{equation} \hat{p}{\theta_i}(y|x) = \frac{1}{T} \sum{t=1}^{T} p_{\theta_i^t}(y|x) \end{equation}

where $p_{\theta_i^t}(y|x)$ represents the softmax output of the $i^{th}$ model at the $t^{th}$ forward pass, each with a different dropout mask. The overall EMCD predictive distribution is then obtained by averaging across all $N$ ensemble members:

\begin{equation} \hat{p}(y|x) = \frac{1}{N} \sum_{i=1}^{N} \hat{p}_{\theta_i}(y|x) \end{equation}

The predictive uncertainty is measured using predictive entropy, defined as:

\begin{equation} PE = -\sum_{i=0}^{C} \hat{p}(y_i|x) \log \hat{p}(y_i|x) \end{equation}

where $C$ denotes the number of output classes. By combining the model diversity from deep ensembles with the stochastic behavior of dropout, EMCD provides a more comprehensive estimation of epistemic uncertainty and can lead to improved model calibration.

\begin{table}[t]
    \caption{Uncertainty confusion matrix}
    \label{Tab:Unc-CF}
    \centering
    \begin{tabular}{cccc}
        \toprule
        & & \multicolumn{2}{c}{Correctness} \\
        & & Correct & Incorrect \\
        \midrule
        \multirow{2}{*}{Confidence} 
        & Certain 
        & \cellcolor[HTML]{ADD8E6} CC 
        & \cellcolor[HTML]{FFB347} IC \\
        & Uncertain 
        & \cellcolor[HTML]{FFB347} CU 
        & \cellcolor[HTML]{ADD8E6} IU \\
        \bottomrule
    \end{tabular}
\end{table}


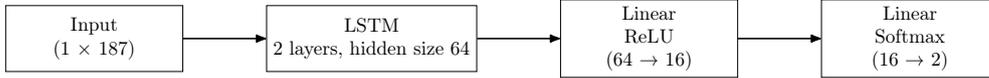
\begin{figure*}[htbp]
    \centering
    \begin{adjustbox}{width=\linewidth}
    \begin{tikzpicture}[
        node distance=2cm and 1.5cm,
        box/.style = {draw, minimum width=3.2cm, minimum height=1.2cm, align=center},
        arrow/.style = {-{Latex[round]}, thick},
        ]

    \node[box] (input) {Input\\(1 × 187)};    
    \node[box, right=of input] (lstm) {LSTM\\2 layers, hidden size 64};    
    \node[box, right=of lstm] (fc1) {Linear\\ReLU\\(64 → 16)};    
    \node[box, right=of fc1] (fc2) {Linear\\Softmax\\(16 → 2)};    

    \draw[arrow] (input) -- (lstm);
    \draw[arrow] (lstm) -- (fc1);
    \draw[arrow] (fc1) -- (fc2);

    \end{tikzpicture}
    \end{adjustbox}
    \caption{Architecture of the LSTM-based model}
    \label{fig:lstm_model_architecture}
\end{figure*}

\begin{figure*}[htbp]
    \centering
    \begin{adjustbox}{width=\linewidth}
    \begin{tikzpicture}[
        node distance=2cm and 1.5cm,
        box/.style = {draw, minimum width=3.2cm, minimum height=1.2cm, align=center},
        arrow/.style = {-{Latex[round]}, thick},
        ]
        
    \node[box] (input) {Input\\(1 × 187)};    
    \node[box, right=of input] (conv1) {Conv1D\\BN\\ReLU\\(1 → 16, K=3, P=2)\\MaxPool1D (K=2)};    
    \node[box, right=of conv1] (conv2) {Conv1D\\BN\\ReLU\\(16 → 32, K=3, P=2)\\MaxPool1D (K=2)}; 
    \node[box, right=of conv2] (fc1) {Linear\\ReLU\\(1536 → 16)};    
    \node[box, right=of fc1] (fc2) {Linear\\Softmax\\(16 → 2)};  

    \draw[arrow] (input) -- (conv1);
    \draw[arrow] (conv1) -- (conv2);
    \draw[arrow] (conv2) -- (fc1);
    \draw[arrow] (fc1) -- (fc2);

    \end{tikzpicture}
    \end{adjustbox}
    \caption{Architecture of the 1D CNN model}
    \label{fig:cnn1d_model_architecture}
\end{figure*}
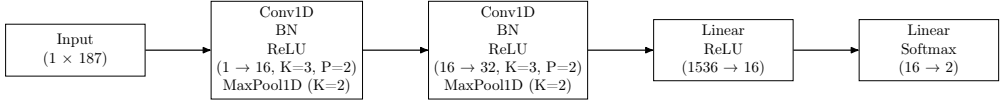

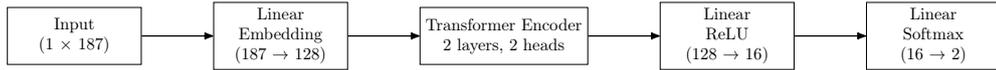
\begin{figure*}[htbp]
    \centering
    \begin{adjustbox}{width=\linewidth}
    \begin{tikzpicture}[
        node distance=2cm and 1.5cm,
        box/.style = {draw, minimum width=2.8cm, minimum height=1.2cm, align=center},
        arrow/.style = {-{Latex[round]}, thick},
        ]

    \node[box] (input) {Input\\(1 × 187)};
    \node[box, right=of input] (embedding) {Linear\\Embedding\\(187 → 128)};
    \node[box, right=of embedding] (transformer) {Transformer Encoder\\2 layers, 2 heads};
    \node[box, right=of transformer] (fc1relu) {Linear\\ReLU\\(128 → 16)};
    \node[box, right=of fc1relu] (fc2softmax) {Linear\\Softmax\\(16 → 2)};

    \draw[arrow] (input) -- (embedding);
    \draw[arrow] (embedding) -- (transformer);
    \draw[arrow] (transformer) -- (fc1relu);
    \draw[arrow] (fc1relu) -- (fc2softmax);

    \end{tikzpicture}
    \end{adjustbox}
    \caption{Architecture of the Transformer model}
    \label{fig:model_architecture}
\end{figure*}
\subsection{Uncertainty Evaluation}
In a classification task, predictions can be divided into two categories: correct and incorrect. In uncertainty-aware classification, each prediction is also associated with a confidence score or certainty level. By plotting the density of certainty values for correct and incorrect predictions, we can generate a visualization similar to Figure \ref{fig:pe_distribution}. Ideally, the density curves for correct and incorrect predictions should be well-separated, with greater separation indicating more effective uncertainty estimation. To quantitatively assess the quality of uncertainty estimation, we employ an uncertainty-aware confusion matrix, which categorizes predictions based on both correctness (correct or incorrect) and confidence (certain or uncertain), as shown in Table~\ref{Tab:Unc-CF} \cite{tabarisaadi2022optimized}. Favorable outcomes include Correct and Certain (CC) and Incorrect and Uncertain (IU), as they reflect either reliable or appropriately cautious predictions. Conversely, Incorrect and Certain (IC) and Correct and Uncertain (CU) are less desirable, particularly IC, which indicates overconfident errors.

\begin{figure*}[t]
    \centering
    \begin{subfigure}{0.48\linewidth}
        \includegraphics[width=\linewidth]{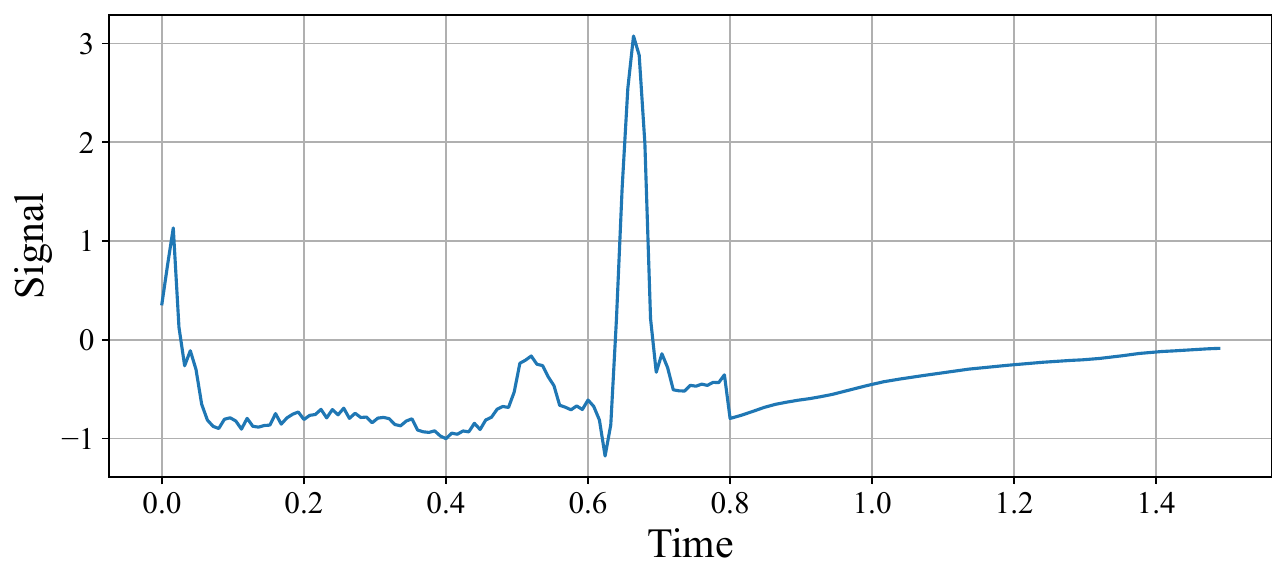}
        \caption{Normal ECG}
        \label{fig:normal_ecg}
    \end{subfigure}
    \vspace{4pt} 
    \begin{subfigure}{0.48\linewidth}
        \includegraphics[width=\linewidth]{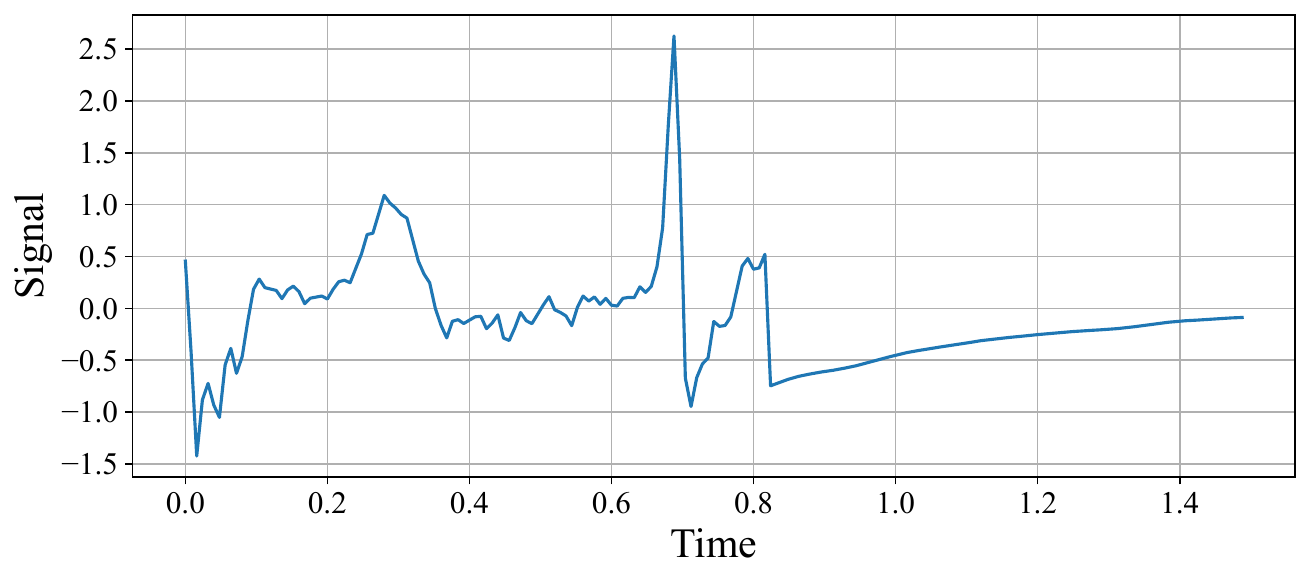}
        \caption{Abnormal ECG}
        \label{fig:abnormal_ecg}
    \end{subfigure}    
    \begin{subfigure}{0.48\linewidth}
        \includegraphics[width=\linewidth]{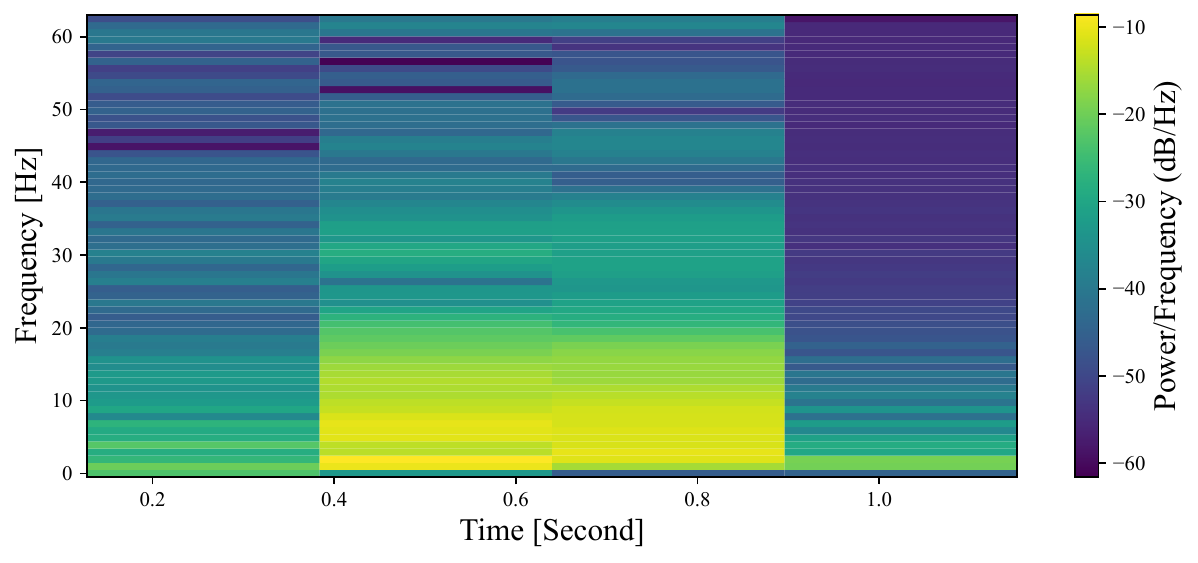}
        \caption{Normal Spectrogram}
        \label{fig:normal_spec}
    \end{subfigure}
    \vspace{4pt} 
    \begin{subfigure}{0.48\linewidth}
        \includegraphics[width=\linewidth]{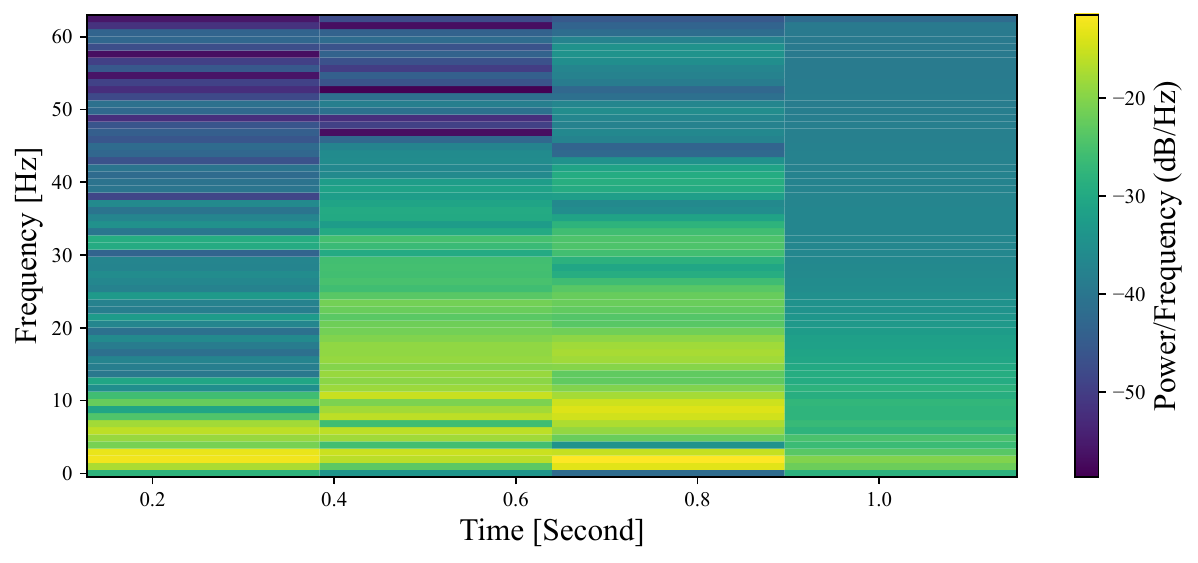}
        \caption{Abnormal Spectrogram}
        \label{fig:abnormal_spec}
    \end{subfigure}   
    \caption{Two samples of normal and abnormal ECG signals and their corresponding spectrograms. Subfigures~\ref{fig:normal_ecg} and \ref{fig:abnormal_ecg} display time-domain ECG waveforms for normal and abnormal cardiac activity, respectively. Subfigures \ref{fig:normal_spec} and \ref{fig:abnormal_spec} illustrate their time–frequency representations, showing clear spectral differences between normal and abnormal patterns.}
    \label{fig:plotsample}
\end{figure*}


Uncertainty Sensitivity measures the proportion of incorrect predictions that are also uncertain, indicating the model’s ability to recognize its own errors:

\begin{equation} U_{Sen} = \frac{N_{\text{IU}}}{N_{\text{IC}} + N_{\text{IU}}} \end{equation}\label{eq9}

Uncertainty Specificity captures the proportion of correct predictions that are also certain, reflecting the model’s confidence in accurate predictions:

\begin{equation} U_{Spe} = \frac{N_{\text{CC}}}{N_{\text{CC}} + N_{\text{CU}}} \end{equation}

Uncertainty Precision
indicates how well uncertainty aligns with incorrect predictions by measuring the proportion of uncertain predictions that are wrong:

\begin{equation} U_{Pre} = \frac{N_{\text{IU}}}{N_{\text{CU}} + N_{\text{IU}}} \end{equation}

Uncertainty Accuracy
represents the overall reliability of uncertainty estimates by quantifying how often the model is either confidently correct or cautiously incorrect:

\begin{equation} U_{Acc} = \frac{N_{\text{CC}} + N_{\text{IU}}}{N_{\text{CC}} + N_{\text{CU}} + N_{\text{IC}} + N_{\text{IU}}} \end{equation}


\begin{figure*}[htbp]
  \centering
  \includegraphics[width=1\linewidth]{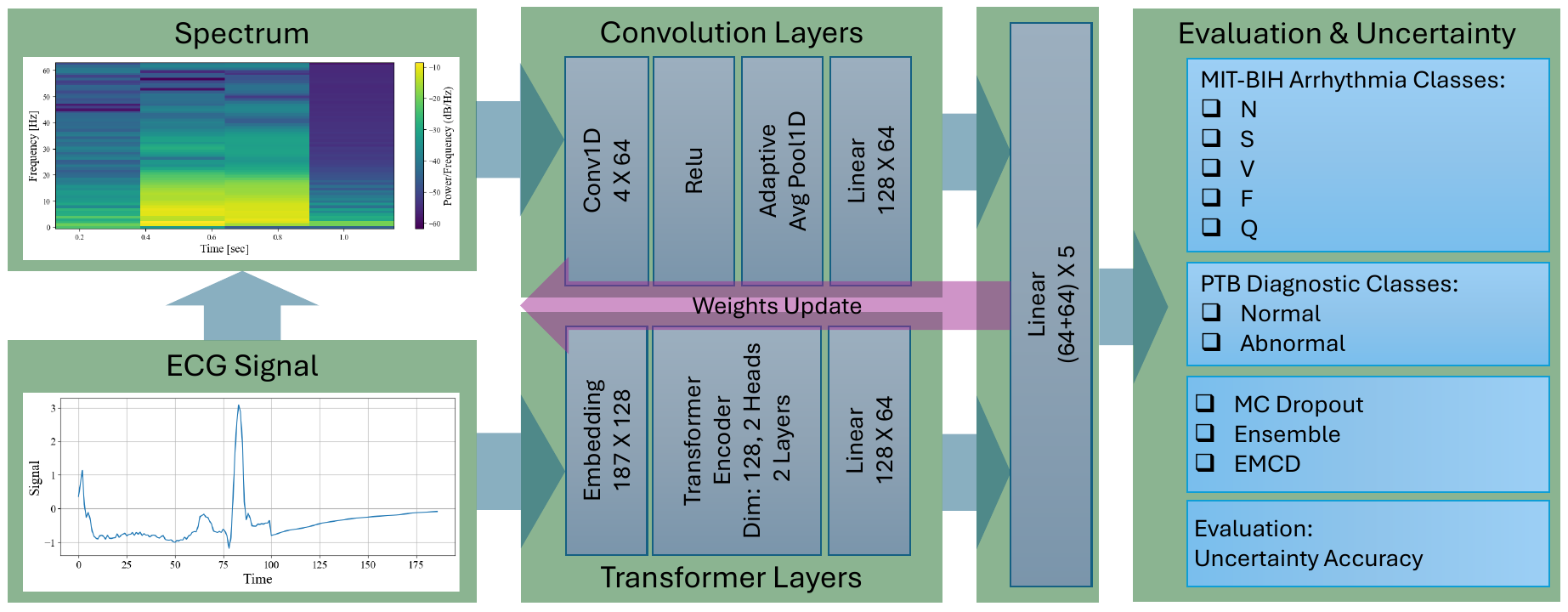} 
    \caption{Architecture of the proposed UCTECG-Net}
    \label{fig:architecture_proposed_method}
\end{figure*}

\subsection{Related studies}
Deep learning techniques have been extensively investigated for automated ECG analysis, with prior studies exploring diverse neural architectures, benchmark datasets, and uncertainty quantification (UQ) strategies. Existing research can generally be grouped into two main directions: (i) works primarily focused on improving classification performance using convolutional, recurrent, or hybrid architectures, and (ii) studies explicitly incorporating uncertainty estimation mechanisms to enhance predictive reliability in safety-critical clinical settings.

Several recent contributions have examined uncertainty-aware learning in multi-label ECG classification and related physiological signal analysis tasks. These include Bayesian neural networks, Monte Carlo dropout, deep ensembles, Dirichlet-based modeling, and subjective logic frameworks. While such approaches demonstrate improved calibration and reliability, many are limited to either single-domain ECG representations or specific diagnostic subsets (e.g., myocardial infarction detection). Moreover, multimodal physiological studies often integrate ECG with other biosignals but do not consistently address uncertainty-aware arrhythmia classification using complementary time and time–frequency representations within a unified architecture.

To provide a structured overview of these developments, Table~\ref{tab:comparison_with_existing_researches} summarizes representative studies in terms of datasets, publication year, modeling approach, evaluation metrics, and whether uncertainty-aware accuracy (UAcc) or related reliability measures were explicitly reported. This comparative synthesis highlights methodological trends, the diversity of evaluation protocols, and existing gaps, particularly the limited integration of multimodal ECG representations with systematic uncertainty quantification across standard arrhythmia benchmarks. These observations motivate the development of the proposed UCTECG-Net framework.
   
\begin{table*}[t]
    \caption{Previous related studies summary}
    \label{tab:comparison_with_existing_researches} 
    \centering
    \resizebox{1\textwidth}{!}{
    \begin{tabular}{llrrrr}
        \toprule
        \makecell[l]{Research} & \makecell[l]{Dataset} & \makecell[l]{Year} 
        & \makecell[l]{Approach} & \makecell[l]{Metrics} & \makecell[l]{UAcc} \\
        \midrule
        
        \makecell[l]{Evaluation of uncertainty\\quantification methods in\\multi-label classification \cite{BARANDAS2024101978}}
        & \makecell[l]{CPSC, G12EC,\\PTB-XL}
        & \makecell[l]{2024}
        & \makecell[l]{BNN-Dropout, BNN-Laplace,\\Deep Ensemble, Bootstrap}
        & \makecell[l]{AUROC, F1 Score,\\ECE}
        & \makecell[l]{Reported} \\ \\
        
        \makecell[l]{Fatigue Detection with Multimodal\\Physiological Signals via\\Uncertainty-Aware Deep\\Transfer Learning \cite{kakhi2026fatigue}}
        & \makecell[l]{Multimodal physiological\\dataset (EEG, ECG,\\EDA, PPG, RESP)}
        & \makecell[l]{2026}
        & \makecell[l]{Deep transfer learning\\MC-Dropout,\\Deep Ensembles, EMCD}
        & \makecell[l]{Accuracy, F1 Score,\\Precision, Recall,\\AUC}
        & \makecell[l]{Reported} \\ \\
        
        \makecell[l]{Uncertainty quantification in\\DenseNet model using myocardial\\infarction ECG signals \cite{JAHMUNAH2023107308}}
        & \makecell[l]{PTB (MI subset)}
        & \makecell[l]{2023}
        & \makecell[l]{Dirichlet\\DenseNet}
        & \makecell[l]{Confusion Matrix,\\Uncertainty\\Confusion Matrix}
        & \makecell[l]{Reported} \\ \\

        \makecell[l]{A Reliable Deep Learning Model for\\ECG Interpretation: Mitigating\\Overconfidence and Direct UQ \cite{Li2025Reliable}}
        & \makecell[l]{PhysioNet\\Challenge 2017}
        & \makecell[l]{2025}
        & \makecell[l]{Residual CNN,\\Dirichlet-based uncertainty,\\Subjective Logic}
        & \makecell[l]{Accuracy, F1 Score,\\PR-AUC, ROC-AUC}
        & \makecell[l]{Not\\reported} \\ \\

        \makecell[l]{A novel inference system for\\detecting cardiac arrhythmia using\\deep learning framework \cite{sai2025novel}}
        & \makecell[l]{MIT-BIH}
        & \makecell[l]{2025}
        & \makecell[l]{DeepBiLSTMnet with\\wavelet sequence layer,\\Bi-LSTM}
        & \makecell[l]{Accuracy, F1 Score,\\Precision, Recall}
        & \makecell[l]{Not\\reported} \\ \\
        
        \makecell[l]{This Research\\UCTECG-Net} 
        & \makecell[l]{MIT-BIH and PTB\\Arrhythmia}
        & \makecell[l]{-}
        & \makecell[l]{Multimodal Spectrum/Signal\\CNN, Transformer\\MC-Dropout,\\Deep Ensembles, EMCD}
        & \makecell[l]{Accuracy, Sensitivity,\\Specificity, F1 Score}
        & \makecell[l]{Reported} \\ \\

        \bottomrule
    \end{tabular}    
    }
\end{table*}


\section{Dataset and Preprocessing}\label{sec:dataset}
\subsection{PhysioNet MIT-BIH Arrhythmia Dataset}

The PhysioNet MIT-BIH Arrhythmia Database \cite{PhysioNet, 932724} is among the most widely used and influential resources in biomedical signal processing research. It comprises 48 half-hour, two-channel ECG recordings collected from 47 subjects, sampled at 360 Hz and carefully annotated by expert cardiologists. The database was established to provide a standardized benchmark for the development, validation, and comparison of automated arrhythmia detection algorithms.

In this study, we utilize the ECG Heartbeat Arrhythmia Dataset \cite{8419425, ecgheartbeat_kaggle} available on Kaggle, which contains a preprocessed subset derived from the MIT-BIH Arrhythmia Database. This dataset comprises segmented heartbeat samples extracted from continuous ECG recordings and annotated with the original expert labels. Each segment comprises 187 time samples, resampled to 125 Hz to ensure uniform sampling.  The dataset contains over one hundred thousand labeled heartbeat segments categorized into five heartbeat classes: Normal (N), Supraventricular (S), Ventricular (V), Fusion (F), and Paced (Q).

Table~\ref{table:datsets}  summarizes the dataset specifications, while Figure~\ref{fig:plotsample} illustrates representative examples of normal and abnormal ECG signals, along with their corresponding spectrograms, highlighting the distinct temporal and spectral characteristics that facilitate effective differentiation between healthy and pathological heartbeats.

\begin{table}[t] 
    \caption{Datasets information}
    \centering
    \begin{tabular}{lllrrr}
    \toprule
        Dataset & MIT-BIH & PTB \\
    \midrule    
    Total Samples & 109,446 & 14,552 \\
    Train Samples & 87,554 (80\%) & 11,641 (80\%) \\
    Total Samples & 21,892 (20\%) & 2,911 (20\%) \\
    Resampled Frequency & 125 Hz & 125 Hz \\
    Time Steps 187 & 187 & 187 \\
    Number of Classes & 5 & 2 \\
    \bottomrule
    \end{tabular}
    \label{table:datsets}
\end{table}

\subsection{PhysioNet PTB Diagnostic Database }

The ECG Heartbeat Categorization Dataset used in this study \cite{8419425, ecgheartbeat_kaggle} is derived from the PhysioNet PTB Diagnostic ECG Database. It comprises 14,552 heartbeat samples categorized into two classes: Normal and Abnormal. Each segment represents a preprocessed heartbeat, downsampled to 125 Hz and standardized to a fixed length of 187 samples via cropping and zero-padding, ensuring consistent input dimensions for model training. 

The original PTB recordings were acquired using a high-fidelity 16-channel ECG recorder with 16-bit resolution and a sampling rate of 1000 Hz, subsequently resampled for uniformity. This dataset offers a reliable and well-structured foundation for developing and evaluating deep learning models aimed at automated heartbeat classification and cardiovascular disease detection.

Table~\ref{table:datsets} summarizes the dataset specifications.
\subsection{Proposed method}
Fig.~\ref{fig:architecture_proposed_method} presents the proposed UCTECG-Net, a hybrid deep learning architecture that integrates convolutional and transformer components to enhance ECG arrhythmia classification with built-in uncertainty awareness. The model jointly processes raw ECG signals and their corresponding time–frequency spectrograms to leverage complementary information from both domains. While raw signals preserve the original temporal morphology of the heartbeat, such as P-wave shape, QRS width, and T-wave dynamics, the spectrogram representation highlights frequency-based patterns that may be less visible in the time domain, including harmonic content, transient events, and frequency shifts associated with specific arrhythmias. Using both views enables the network to capture richer discriminative features and improves robustness against noise and inter-patient variability.

The convolutional branch processes the raw ECG waveform through four 1D convolutional layers with 64 filters each, followed by ReLU activations and adaptive average pooling. This branch focuses on extracting localized patterns and morphological cues from the signal. In parallel, the transformer branch embeds the input into a 128-dimensional latent space. It is processed by a two-layer Transformer encoder with two attention heads, enabling the extraction of long-range dependencies and global contextual relationships across the heartbeat sequence.

The feature representations from both branches are fused and passed through fully connected layers to generate class probabilities for arrhythmia detection. UCTECG-Net is designed to support multi-class classification on the MIT-BIH dataset (N, S, V, F, Q) and binary classification (Normal vs. Abnormal) on the PTB dataset.

To enhance model reliability, we incorporate three uncertainty quantification approaches, MCD, Deep Ensembles, and EMCD, within the evaluation pipeline. These methods enable the model to express confidence in its predictions, helping clinicians identify ambiguous or uncertain cases and reducing the risk of misclassification. Overall, UCTECG-Net is optimized for uncertainty-aware performance and is well-suited to high-stakes clinical applications in which trust, interpretability, and reliability are essential.

\subsection{Other models Architecture}
To classify ECG signals into normal and abnormal categories, we designed and evaluated three deep learning architectures: an LSTM network, a one-dimensional CNN1D, and a Transformer-based model. Each architecture performs binary classification and outputs class probabilities for the two heartbeat categories.

The LSTM model (Figure~\ref{fig:lstm_model_architecture}) consists of two stacked LSTM layers with a hidden size of 64, followed by fully connected layers with 16 and 2 units. ReLU activation and dropout (0.2) are applied between layers, and a softmax function is used in the final layer to generate class probabilities.

The CNN1D model (Figure~\ref{fig:cnn1d_model_architecture}) contains two convolutional blocks, with 16 filters in the first block and 32 in the second. Each block includes batch normalization, ReLU activation, and max pooling. The resulting feature maps are flattened and passed through two fully connected layers, with ReLU and dropout applied before the softmax output layer.

The Transformer model (Figure~\ref{fig:model_architecture}) begins with a linear embedding layer that projects each 187-sample input segment into a 128-dimensional representation. This is followed by a Transformer encoder comprising two layers, each with two attention heads. The encoder outputs are averaged over the temporal dimension and subsequently passed through two fully connected layers with ReLU activation and dropout, followed by a softmax layer for final classification.

\section{Simulation and Results}\label{Simulation and Results}

\begin{table*}[t] 
    \centering
    \caption{Comparison of classification performance for Transformer, CNN1D, and LSTM models vs proposed UCTECG-Net. Results are reported as mean ± standard deviation over 5 runs.}
    \label{table:acc_table}
    \begin{tabular}{llllllrrrrrr}
    \toprule
        Model & Model & Accuracy & Precision & Recall & F1 Score\\ 
    \midrule 
        & LSTM & 96.16 ± 2.92 & 95.57 ± 3.74 & 96.16 ± 2.92 & 95.56 ± 3.74 \\
        MIT-BIH & CNN1D & 97.87 ± 0.08 & 97.79 ± 0.09 & 97.87 ± 0.08 & 97.77 ± 0.08 \\
        & Transformer & 98.53 ± 0.02 & 98.50 ± 0.02 & 98.53 ± 0.02 & 98.49 ± 0.02 \\
        & UCTECG-Net & \textbf{98.58} ± 0.02 & \textbf{98.54} ± 0.02 & \textbf{98.58} ± 0.02 & \textbf{98.54} ± 0.02 \\
        \midrule
        & LSTM & 79.57 ± 4.12 & 78.91 ± 5.11 & 79.57 ± 4.12 & 78.16 ± 6.83 \\
        PTB & CNN1D & 98.65 ± 0.32 & 98.65 ± 0.32 & 98.65 ± 0.32 & 98.65 ± 0.32 \\
        & Transformer & 98.40 ± 0.13 & 98.40 ± 0.13 & 98.40 ± 0.13 & 98.40 ± 0.13 \\
        & UCTECG-Net & \textbf{99.14} ± 0.14 & \textbf{99.14} ± 0.14 & \textbf{99.14} ± 0.14 & \textbf{99.14} ± 0.14 \\
    \bottomrule
    \end{tabular}
\end{table*}


\begin{figure*}[t]
    \begin{subfigure}{0.49\linewidth}
        \centering
        \includegraphics[width=0.8\linewidth]{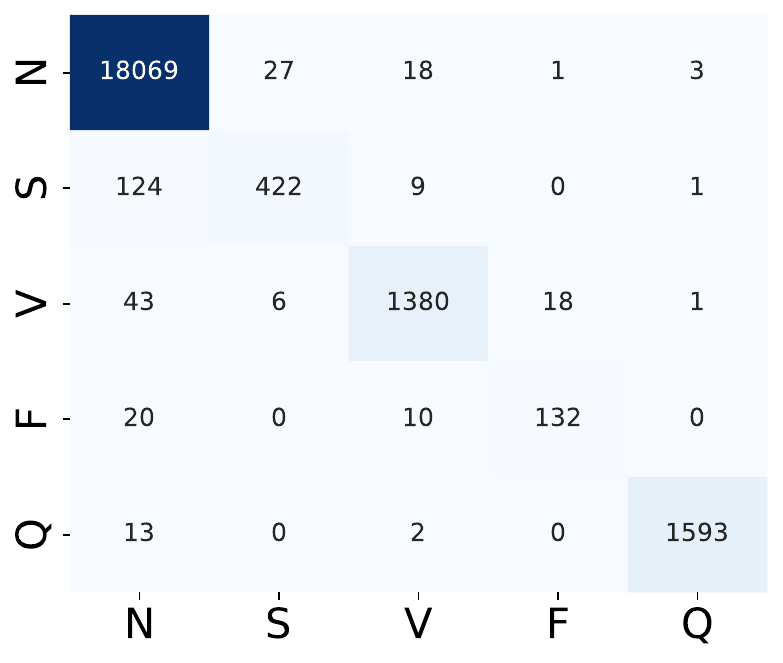}
        \caption{MIT-BIH}
        \label{fig:mitbih_UCTECG-Net_cm}
    \end{subfigure}
    \hfill
    \begin{subfigure}{0.49\linewidth}
        \centering
        \includegraphics[width=0.8\linewidth]{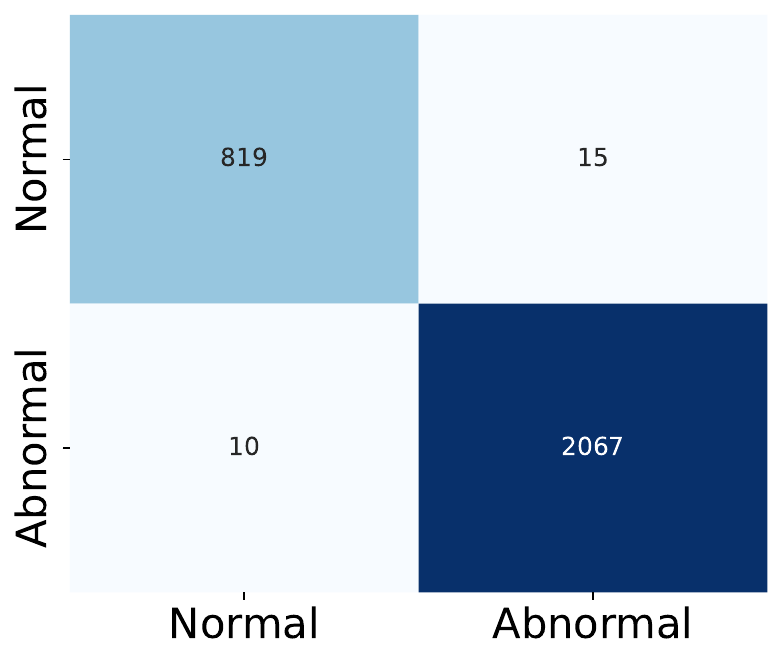}
        \caption{PTB}
        \label{fig:ptbdb_UCTECG-Net_cm}
    \end{subfigure}   
    \caption{Confusion matrices of the best model, UCTECG-Net, for MIT-BIH (Subfigure~\ref{fig:mitbih_UCTECG-Net_cm}) and PTB (Subfigure~\ref{fig:ptbdb_UCTECG-Net_cm}).}
    \label{fig:UCTECG-Net_cm}
\end{figure*}


\subsection{Performance Evaluation}
To ensure a fair and reliable comparison, each model (LSTM, CNN1D, Transformer, and the proposed UCTECG-Net) was independently trained and evaluated over five runs with different random initializations. As reported in Table~\ref{table:acc_table}, UCTECG-Net consistently achieved the highest overall performance across both datasets. On the MIT-BIH Arrhythmia dataset, the proposed model achieved an accuracy of 98.58 ± 0.02\%, precision of 98.54 ± 0.02\%, recall of 98.58 ± 0.02\%, and F1 score of 98.54 ± 0.02\%, surpassing the strong baseline provided by the transformer model. A similar trend was observed on the PTB dataset, where UCTECG-Net reached an accuracy of 99.14 ± 0.14\%, demonstrating reliable generalization across distinct ECG classification tasks with varying signal characteristics.

The transformer model delivered competitive results, 98.53 ± 0.02\% accuracy on MIT-BIH and 98.40 ± 0.13\% on PTB. In contrast, the LSTM model showed the weakest and most variable performance, likely due to its limited ability to extract discriminative high-frequency and morphological features from complex ECG patterns.

The superior performance of UCTECG-Net can be attributed to its hybrid design, which integrates convolutional layers to capture localized morphological cues (e.g., QRS shape, P-wave distortions) and transformer layers to model long-range temporal dependencies and cross-beat contextual patterns. This dual-branch architecture enables the network to learn a richer, more complementary representation of the ECG signal than any single-branch model. Furthermore, the inclusion of frequency-domain information through spectrograms enhances the model’s ability to detect subtle abnormalities that may be less prominent in the raw time-domain waveform. The combined effect of multi-resolution temporal–spectral analysis and hybrid feature extraction leads to improved classification accuracy, greater stability across runs, and better handling of inter-patient variability, particularly on challenging datasets such as PTB.

The confusion matrices of UCTECG-Net, shown in Fig. \ref{fig:mitbih_UCTECG-Net_cm} for MIT-BIH and Fig. \ref{fig:ptbdb_UCTECG-Net_cm} for PTB, further illustrate the robustness of the proposed model. For the PTB dataset, the binary confusion matrix indicates a very low misclassification rate, with 819 normal and 2067 abnormal beats correctly identified, and only a small number of false positives and false negatives. This balanced error distribution confirms the model’s high sensitivity and specificity, demonstrating its ability to reliably distinguish pathological ECG signals from normal ones without bias toward a particular class 

On the MIT-BIH dataset, the multi-class confusion matrix shows strong diagonal dominance across all heartbeat categories (N, S, V, F, and Q), indicating accurate class-wise discrimination. The normal (N) and ventricular (V) classes are detected with particularly high confidence, while misclassifications between morphologically similar classes (e.g., S and V) remain minimal. Importantly, rare classes such as F and Q are also well recognized, reflecting the model’s ability to handle class imbalance and subtle morphological variations. These results confirm that UCTECG-Net not only achieves high aggregate performance metrics but also maintains consistent and clinically meaningful predictions across individual heartbeat types

\begin{table*}[t]
    \centering
    \caption{Comparison of uncertainty quantification results for the Transformer model using MCD, Ensemble, and EMCD methods.}
    \label{table:uacc_combined}
    \begin{tabular}{lllllllllrrrrrrrrrr}
    \toprule
        Dataset & Model & Method & CU & IC & IU & CC & UAcc & USen & USpe & UPre \\
    \midrule
        & LSTM & MCD & 998 & 926 & 2719 & 17249 & 83.35 & 51.87 & 86.38 & 26.85 \\
        & LSTM & Ensemble & 315 & 142 & 1862 & 19573 & 90.85 & 68.93 & 91.31 & 14.47 \\
        & LSTM & EMCD & 329 & 128 & 2012 & 19423 & 90.22 & 71.99 & 90.61 & 14.05 \\
        & CNN1D & MCD & 439 & 205 & 1139 & 20109 & 93.86 & 68.17 & 94.64 & 27.82 \\
        MIT-BIH & CNN1D & Ensemble & 190 & 267 & 293 & 21142 & 97.44 & 41.58 & 98.63 & 39.34 \\
        & CNN1D & EMCD & 434 & 224 & 870 & 20364 & 95.0 & 65.96 & 95.9 & 33.28 \\
        & Transformer & MCD & 125 & 205 & 273 & 21289 & 97.82 & 37.88 & 98.73 & 31.41 \\
        & Transformer & Ensemble & 88 & 205 & 151 & 21448 & 98.37 & 30.03 & 99.3 & 36.82 \\
        & Transformer & EMCD & 113 & 181 & 219 & 21379 & 98.17 & 38.44 & 98.99 & 34.04 \\
        & UCTECG-Net & MCD & 71 & 237 & 119 & 21465 & 98.37 & 23.05 & 99.45 & 37.37 \\
        & UCTECG-Net & Ensemble & 110 & 186 & 141 & 21455 & \textbf{98.51} & 37.16 & 99.35 & 43.82 \\
        & UCTECG-Net & EMCD & 120 & 178 & 168 & 21426 & 98.42 & 40.27 & 99.22 & 41.67 \\
        \midrule
        & LSTM & MCD & 417 & 206 & 739 & 1549 & 67.54 & 66.93 & 67.7 & 36.07 \\
        & LSTM & Ensemble & 344 & 142 & 960 & 1465 & 62.14 & 70.78 & 60.41 & 26.38 \\
        & LSTM & EMCD & 345 & 140 & 965 & 1461 & 62.04 & 71.13 & 60.22 & 26.34 \\
        & CNN1D & MCD & 29 & 26 & 84 & 2772 & 96.22 & 52.73 & 97.06 & 25.66 \\
        PTB & CNN1D & Ensemble & 11 & 23 & 46 & 2831 & 97.63 & 32.35 & 98.4 & 19.3 \\
        & CNN1D & EMCD & 16 & 22 & 45 & 2828 & 97.7 & 42.11 & 98.43 & 26.23 \\
        & Transformer & MCD & 22 & 34 & 29 & 2826 & 97.84 & 39.29 & 98.98 & 43.14 \\
        & Transformer & Ensemble & 18 & 25 & 37 & 2831 & 97.87 & 41.86 & 98.71 & 32.73 \\
        & Transformer & EMCD & 20 & 22 & 42 & 2827 & 97.8 & 47.62 & 98.54 & 32.26 \\
        & UCTECG-Net & MCD & 8 & 19 & 26 & 2858 & 98.45 & 29.63 & 99.1 & 23.53 \\
        & UCTECG-Net & Ensemble & 9 & 16 & 11 & 2875 & 99.07 & 36.0 & 99.62 & 45.0 \\
        & UCTECG-Net & EMCD & 11 & 16 & 9 & 2875 & \textbf{99.14} & 40.74 & 99.69 & 55.0 \\
    \bottomrule
    \end{tabular}
\end{table*}
\subsection{Uncertainty Quantification Results}
To comprehensively assess predictive uncertainty, three widely used UQ techniques (MCD, Deep Ensembles, and EMCD) were implemented across all baseline architectures and the proposed UCTECG-Net. The combined results in Table~\ref{table:uacc_combined} demonstrate that, while all approaches retained high classification performance, clear differences emerged in their ability to produce reliable uncertainty estimates.

Across both the MIT-BIH and PTB datasets, UCTECG-Net consistently delivered higher uncertainty-aware performance compared to the baseline Transformer, CNN1D, and LSTM models. On the MIT-BIH dataset, UCTECG-Net with the Ensemble method achieved a UAcc of 98.51\%, outperforming the Transformer’s best UAcc of 98.37\%. Similarly, on the PTB dataset, UCTECG-Net achieved a UAcc of 99.14\% with EMCD and 99.07\% with Ensembles, surpassing all Transformer-based configurations. These improvements were also reflected in uncertainty sensitivity, uncertainty specificity, and uncertainty precision, highlighting UCTECG-Net’s enhanced ability to discriminate between correctly, incorrectly, and uncertainly predicted samples.

The superior uncertainty performance of UCTECG-Net can be attributed to several architectural advantages. First, the hybrid representation, which combines convolutional layers and transformer encoders, produces more expressive and stable latent features, thereby strengthening the separation between confident and uncertain regions in the feature space. Convolutional layers effectively capture local morphological cues, while transformer layers encode global dependencies, enabling the model to form more coherent uncertainty boundaries. Second, the dual-branch input (time- and frequency-domain information) provides richer signal diversity, enabling uncertainty estimation methods to more accurately identify cases in which spectral or temporal cues are ambiguous. Finally, UCTECG-Net exhibits lower variance across multiple runs, allowing ensemble-based methods to generate more reliable epistemic uncertainty estimates. Collectively, these factors enable UCTECG-Net not only to achieve higher accuracy but also to offer more trustworthy and calibrated uncertainty estimates, an essential requirement for safety-critical ECG applications.
\begin{figure*}[t]
    \centering
    \begin{subfigure}{\linewidth}
        \includegraphics[width=\textwidth]{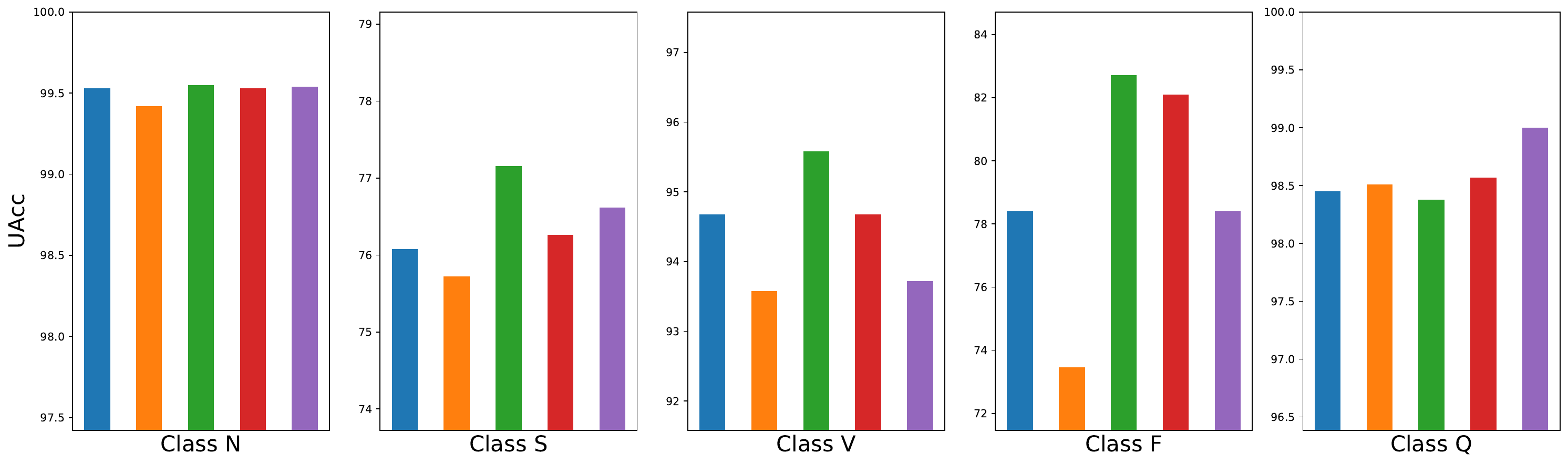}
        \caption{MIT-BIH}
        \label{fig:mitbih_uacc_perclass}
    \end{subfigure}
    \vspace{4pt} 
    \begin{subfigure}{\linewidth}
        \includegraphics[width=\columnwidth]{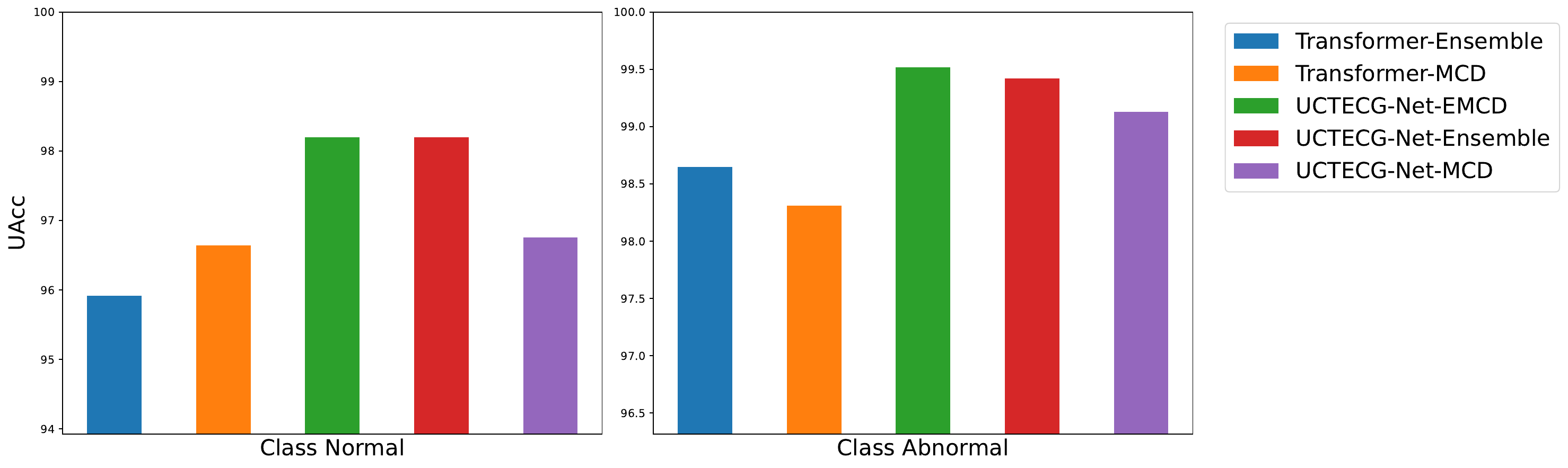}
        \caption{PTB}
        \label{fig:ptbdb_uacc_perclass}
    \end{subfigure}   
    \caption{Subfigures~\ref{fig:mitbih_uacc_perclass} shows Uncertainty Accuracy per class for MIT-BIH. Subfigures \ref{fig:ptbdb_uacc_perclass} shows Uncertainty Accuracy per class for PTB}
    \label{fig:uacc_perclass}
\end{figure*}

\subsection{Scope and Future Directions}
The proposed UCTECG-Net was evaluated on two widely used public ECG datasets, enabling reproducible benchmarking and fair comparison with existing methods. Still, it does not fully capture the variability of real-world clinical settings, such as multi-lead recordings and diverse patient populations. The hybrid convolution–transformer design prioritizes robust feature learning and reliable uncertainty estimation, at the expense of increased computational cost compared to simpler architectures, making the current implementation most suitable for offline analysis or moderately resourced systems. Uncertainty estimation was assessed using established data-driven metrics rather than clinical decision thresholds, and classification was performed on short heartbeat segments rather than on continuous ECG streams, thereby facilitating controlled evaluation while leaving real-time deployment as a natural extension. Future work will focus on extending the framework to multi-lead and long-duration ECG signals, incorporating clinician-in-the-loop calibration, developing lightweight variants for edge deployment, and exploring additional uncertainty quantification and interpretability techniques.

\section{Conclusion}\label{conclusion}
In this study, we introduced UCTECG-Net, a hybrid convolution–transformer architecture designed to enhance both the accuracy and reliability of ECG heartbeat classification. By combining raw ECG signals with their corresponding spectrogram representations, the model captures complementary temporal and time–frequency features, enabling more robust and discriminative learning. Across both the MIT-BIH and PTB datasets, UCTECG-Net consistently outperformed traditional LSTM, CNN1D, and Transformer models, demonstrating superior accuracy, uncertainty accuracy, and overall stability.
A key contribution of this work was the integration of uncertainty quantification into the evaluation pipeline. By applying Monte Carlo Dropout, Deep Ensembles, and EMCD, we assessed not only classification performance but also the trustworthiness of the model’s confidence estimates. The results showed that UCTECG-Net, especially when paired with Ensemble or EMCD, produces more reliable uncertainty estimates that better distinguish confident correct predictions from ambiguous or potentially incorrect ones. This makes the model more suitable for high-stakes clinical environments where transparency and risk-awareness are critical. Overall, the findings demonstrate that combining hybrid feature extraction with uncertainty-aware evaluation yields more reliable ECG classification models. Future work may extend the framework to multi-lead ECGs, longer recordings, or real-world clinical scenarios to further strengthen its applicability and clinical impact.


\section{Declaration}

\subsection{Ethical Approval and Consent to Participate}
Not applicable.

\subsection{Consent for Publication}
Not applicable.

\subsection{Availability of data and material}
Not applicable.

\subsection{Funding}
Not applicable.

\subsection{Conflict of interest}
On behalf of all authors, the corresponding author states that there is no conflict of interest.

\subsection{Authors' contributions}
Hamzeh Asgharnezhad and Pegah Tabarisaadi performed the simulations, prepared illustrations, and drafted the manuscript. Abbas Khosravi, Roohallah Alizadehsani, and U. Rajendra Acharya supervised the study, contributed to its design, and provided critical revisions. All authors approved the final manuscript..


\bibliography{main.bib}

\end{document}